\documentclass{article}
\usepackage{spconf,amsmath,epsfig}
\usepackage{spconf,amsmath,graphicx}
\usepackage{amsmath}
\usepackage{amssymb}
\usepackage{float}
\usepackage{multirow}
\usepackage{longtable}
\usepackage{booktabs}

\usepackage{color}
\usepackage{pifont}

\newcommand{\cmark}{\ding{51}}
\newcommand{\xmark}{\ding{55}}
\def\x{{\mathbf x}}
\def\L{{\cal L}}

\let\OLDthebibliography\thebibliography
\renewcommand\thebibliography[1]{
  \OLDthebibliography{#1}
  \setlength{\parskip}{0pt}
  \setlength{\itemsep}{0pt plus 0.3ex}
}

\begin{document}\sloppy

\def\x{{\mathbf x}}
\def\L{{\cal L}}

\title{Video sentence grounding with temporally global textual knowledge}
%
\name{$\textit{Chen Cai}^1$, $\textit{Runzhong Zhang}^1$, $\textit{Jianjun Gao}^1$, $\textit{Kejun Wu}^1$, $\textit{Kim-Hui Yap}^1$, $\textit{Yi Wang}^2$}
\address{$^1$Nanyang Technological University, $^2$The Hong Kong Polytechnic University}



\maketitle

\begin{abstract}
Temporal sentence grounding involves the retrieval of a video moment with a natural language query. Many existing works directly incorporate the given video and temporally localized query for temporal grounding, overlooking the inherent domain gap between different modalities. 
In this paper, we utilize pseudo-query features containing extensive temporally global textual knowledge sourced from the same video-query pair, to enhance the bridging of domain gaps and attain a heightened level of similarity between multi-modal features.
Specifically, we propose a Pseudo-query Intermediary Network (PIN) to achieve an improved alignment of visual and comprehensive pseudo-query features within the feature space through contrastive learning.
Subsequently, we utilize learnable prompts to encapsulate the knowledge of pseudo-queries, propagating them into the textual encoder and multi-modal fusion module, further enhancing the feature alignment between visual and language for better temporal grounding.
Extensive experiments conducted on the Charades-STA and ActivityNet-Captions datasets demonstrate the effectiveness of our method.
\end{abstract}
\section{Introduction}
\label{sec:intro}

\begin{figure}[!t]
\centering
\includegraphics[width=3.3in]{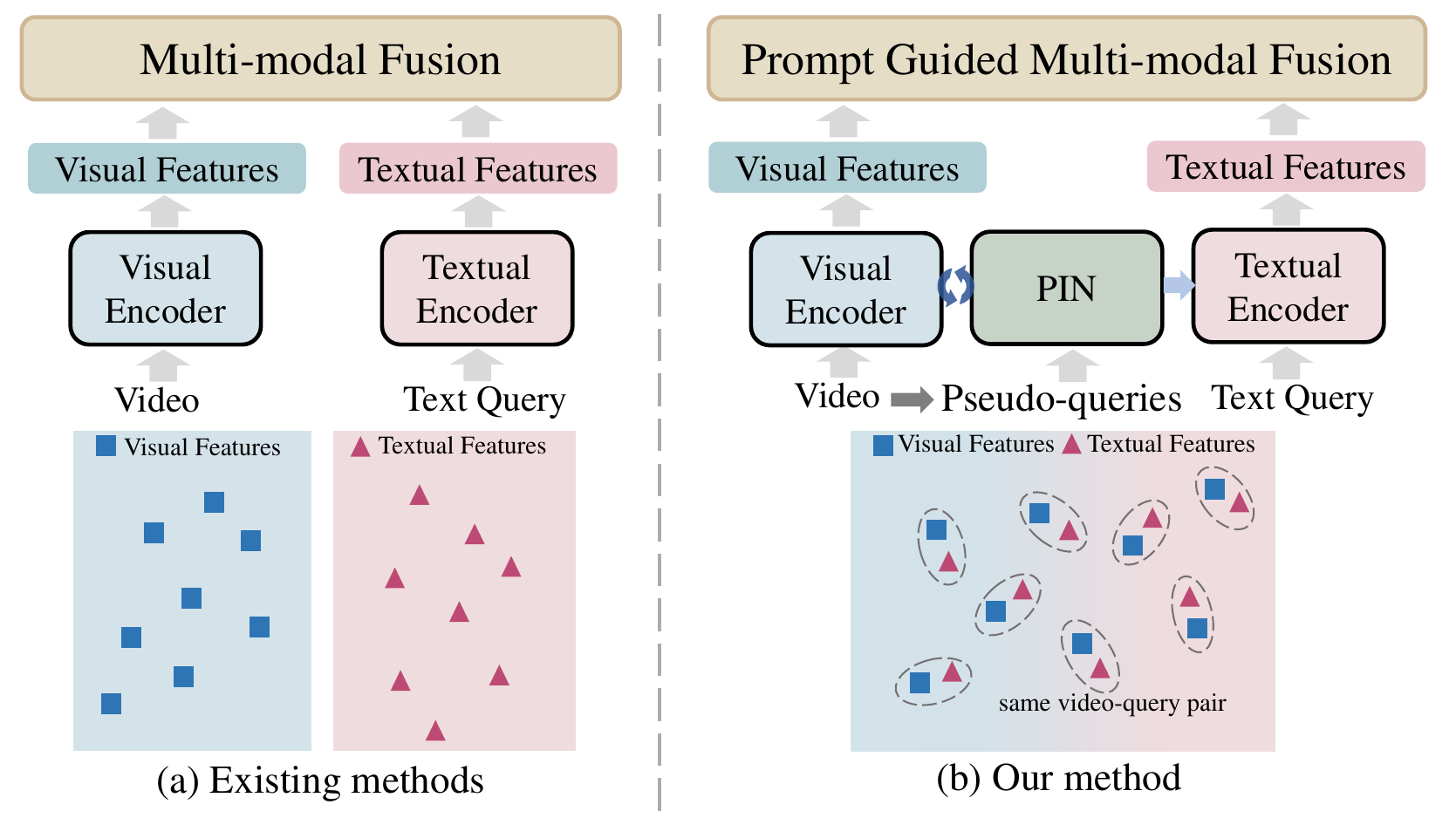}
\caption{(a) Most existing methods \cite{2DTAN, HCL, VSLNet, TCFN} directly integrate the given video and  temporally localized query features, overlooking the inherent domain gaps between different modalities. (b) We introduce a PIN network to further bridge domain gaps, incorporating additional comprehensive temporally global textual knowledge that enhances the overall similarity between multi-modal features. In addition, we utilize the learnable prompt to enhance the alignment learning of the multi-modal fusion module. }
\label{fig_1}
\end{figure}

Temporal sentence grounding (TSG) aims to locate the video segment that aligns semantically with the given natural language query. TSG blurs the boundary between natural language processing and computer vision, establishing a profound foundation for interactive cross-modal applications \cite{9356448}\cite{lan2023survey}. In light of the intrinsic correlation between visual and textual information, recent approaches \cite{Man, PGA, CPN, GIC} have proposed various multi-modal models for the effective fusion of video and textual query features. However, the primary challenge in achieving precise temporal alignment remains in bridging the domain gap between the distinct representations of visual and textual modalities.

To enhance the alignment between video segment and textual query, existing approaches \cite{PGA, AMR, VSLNet} generally adopt ``from visual and textual encoder to multi-modal fusion" method. When presented with a video-query pair as input, the visual and textual encoders initially transform the video frames and temporally localized query words independently into respective feature spaces (shown in Fig.1). Subsequently, the fusion network incorporates the interaction between visual and textual encoders' outputs, generating the multi-modal representation \cite{zhang2023temporal}. However, to learn this representation, most existing methods typically deploy either element-wise multiplication \cite{AMR} or a cross-attention network on top of the features \cite{VSLNet, EMB}. 
In this case, the multi-modal fusion may encounter difficulties in capturing the correlation between textual and visual outputs due to the dissimilarity of features originating from different modalities.
Despite extensive efforts dedicated to the design of multi-modal fusion, the performance remains unsatisfactory due to the inherent domain gap between the visual and temporally localized textual features.

\begin{figure*}[!t]
\centering
\includegraphics[width=7in]{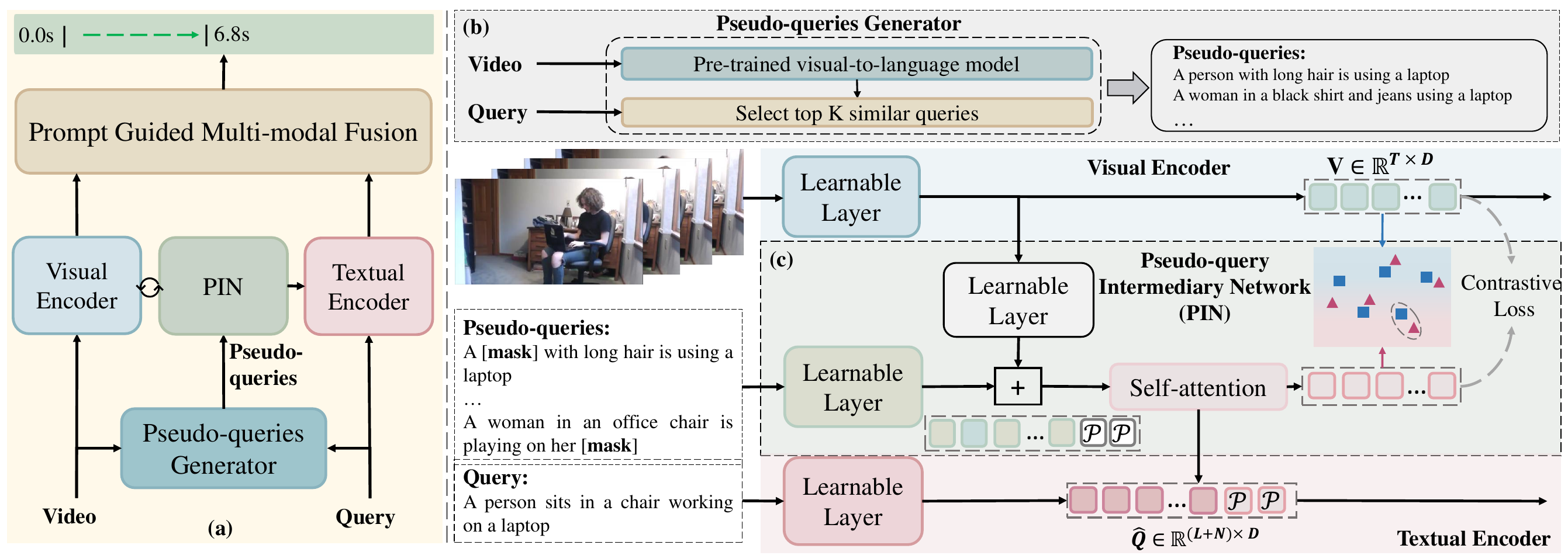}
\caption{ (\textbf{a}) an overview of our proposed method. 
(\textbf{b}) illustrated the comprehensive pseudo-queries generation process based on the untrimmed video and language query. 
Subsequently, as illustrated in \textbf{c}, we employ a Pseudo-query Intermediary Network (PIN) to effectively bridge the multi-modal domain gap by contrastively learning the visual and pseudo-query features to improve the similarity of visual and language features.
Furthermore, the learnable PQ-prompt $ \mathbf{\mathcal{P}}$ outputs from the self-attention layer, which encompasses global textual knowledge, are integrated into the textual encoder to enhance feature alignment in the multi-modal fusion module (shown in Figure \ref{fig_3}).
At the inference stage, pseudo-queries are replaced by the ground truth query to retrieve the relevant PQ-prompt from the prompt pool, enhancing the target moment prediction.
}
\label{fig_2}
\vspace{-6pt}
\end{figure*}

In this paper, we utilize pseudo-query features that contain rich temporally global textual knowledge across the entire video to bridge the domain gap between video and language modalities. We introduce  \textbf{P}seudo-query \textbf{I}ntermediary \textbf{N}etwork named \textbf{PIN} to align the mutual knowledge between visual and textual features. Specifically, we contrastively train the visual encoder with the pseudo-query intermediary network, such that the visual feature and pseudo-query textual feature originating from the same video-query pair exhibit a higher degree of similarity (as illustrated in Fig.1 (b)). 
Furthermore, we introduce the learnable PQ-prompt that encapsulates knowledge of pseudo-query global textual features into the prompt pool, and refine the feature output of the textual encoder to improve the learning of the multi-modal fusion module. 
PQ-prompt contains rich global textual knowledge which is able to enhance cross-modal alignment learning in the multi-modal fusion module.
During the inference stage, we eliminate the need for pseudo-queries, replacing them with ground truth temporally localized queries to retrieve the relevant PQ-prompt from the prompt pool, guiding the model for better temporal grounding.


Our key contributions are summarized as: (1) We propose to bridge the domain gap between multi-modal features by leveraging rich temporally global textual knowledge that is highly relevant to the video and temporally localized text queries. (2) We introduce the Pseudo-query Intermediary Network (PIN) to contrastively align visual features with temporally global textual knowledge to enhance the similarity between visual and language features. Furthermore, we leverage the PQ-prompt to propagate the knowledge, enhancing the learning of feature alignment within the multi-modal fusion module. (3) We validate the proposed method on the Charades-STA and ActivityNet-Captions datasets, achieving state-of-the-art results on most evaluation metrics.

\section{Related work}
\label{sec:format}
 \textbf{Temporal sentence grounding.}
 To address the temporal sentence grounding task and accurately predict moment boundaries, existing methods \cite{2DTAN, VSLNet, HCL, EMB, IVG, ACC, PSL} typically follow the process of generating visual and textual features from pretraining encoders and then designing multi-modal fusion models to align the two modalities. 
 Zhang et al. \cite{2DTAN} introduced a novel two-dimensional temporal map to model the temporal adjacent relations of the video moment.
 Zhang et al. \cite{HCL} propose hierarchical contrastive learning for multi-modality alignment.
 Most methods focused on aligning the visual and temporally localized textual features. In contrast, we explore using temporally global textual knowledge across the entire video to bridge the inherent domain gap of visual and text modalities.

\noindent\textbf{Prompt for transfer learning.} The high-level idea of prompting \cite{ProCL, textPro} is to modify the input features, providing the model with additional information. Recently, prompting has been adopted in many research works, such as prompt tuning \cite{textPro}, visual prompting \cite{visualPro}, and prompt for continuous learning \cite{ProCL}. In this paper, we explore the use of the prompt strategy to enhance multi-modality alignment in temporal sentence grounding.

\section{Methodology}
\label{sec:format}
Given the untrimmed video $\mathcal{V}$ and language query $\mathcal{Q}$, the TSG aims to identify the temporal boundary of the target moment ($\tau_s$,$\tau_e$) within the video $\{v_t\}_{t=\tau_s}^{\tau_e}$ with given $\mathcal{Q}$, where $\tau_s$ and $\tau_e$ denotes start and end of the moment interval. 
The visual pre-trained network is adopted to extract the visual features from $\mathcal{V}$ to $\mathbf{V} = \{\mathbf{v}_t\}_{t=1}^T \in \mathbb{R}^{T \times D}$, and query $\mathcal{Q}$ is encoded to query features $\mathbf{Q} = \{\mathbf{q}_l\}_{l=1}^L \in \mathbb{R}^{L\times D}$, where two independent learnable projection layers are utilized to projection the multi-modal features in same dimension $D$, $T$ and $L$ denotes the number of frames in the video and length of query, respectively. Most of the existing methods \cite{VSLNet, EMB, HCL} encode the visual and query features with attention mechanism \cite{Transformer}:
\begin{align}
    \Bar{\mathbf{V}} = \text{VisualEncoder}(\mathbf{V}), \
     \Bar{\mathbf{Q}} = \text{TextualEncoder}(\mathbf{Q})
\end{align}
furthermore, calculate the similarity score $\mathcal{S} = \text{Softmax}(\text{FC}(\Bar{\mathbf{V}})^\top\text{FC}(\Bar{\mathbf{Q}}))$ and model their semantic correlations in the multi-modal fusion module:
\begin{align}
    \mathbf{V}^{\prime} = \Bar{\mathbf{V}} + \text{FC}(\Bar{\mathbf{V}})\mathcal{S}, \
    \mathbf{Q}^{\prime} = \Bar{\mathbf{Q}} + \text{FC}(\Bar{\mathbf{Q}})\mathcal{S}, \
\end{align}
Finally, the multi-modal representations $\mathbf{V}^{\prime}$ and $\mathbf{Q}^{\prime}$ are fused and projected to predict the start $\tau_s$ and end $\tau_e$ target moment. 
The existing methods directly integrate the multi-modal representation, which overlooks the underlying domain gap inherent between $\mathbf{V}$ and temporally localized $\mathbf{Q}$. 
This oversight can lead to the computation of inferior similarity scores ($\mathcal{S}$) and predicts suboptimal target moments.

As shown in Fig.2, we generate temporally global pseudo-queries to assist in bridging the multi-modal domain gap between $\mathbf{V}$ and $\mathbf{Q}$  using proposed \textbf{P}seudo-query \textbf{I}ntermediary \textbf{N}etwork (\textbf{PIN}).
\subsection{Temporally Global Pseudo-query Generation}
We generate temporally global pseudo-queries from video frames leveraging an advanced pre-trained vision-to-language model (e.g., BLIP2 \cite{blip2}). We uniformly sample $F$ frames $\{v_f\}_{f=1}^F$ from raw video $\mathcal{V}$, and generate frame-specific pseudo-queries $\{P_f\}_{f=1}^F$. We then filter the frame-specific pseudo-queries by selecting the Top-K similar pseudo-queries that have higher cosine similarity scores with respect to the ground truth query. This selection process ensures that the chosen pseudo-queries are highly relevant to the respective video moment. 
Specifically, we adopt the pre-trained language model (e.g., Glove \cite{glove}, GPT2 \cite{GPT2}) to encode the ground truth query as $\mathbf{Q} \in \mathbb{R}^{L\times D}$ and each frame-specific pseudo-query as $ \{\mathbf{P}_f\}_{f=1}^F \in \mathbb{R}^{L\times D}$ to measure the similarity score of $\mathcal{S}_c$ between each of them. The temporally global pseudo-queries $P =\{P_g\}_{g=1}^G \in \mathbb{R}^{L_g\times D} $ with Top-K score in $\mathcal{S}_c$ are chosen to effectively bridge the domain gap with \textbf{PIN}.

\subsection{Pseudo-query Intermediary Network}
The PIN network integrates extensive temporally global textual knowledge in addition to a temporally localized query, improving the similarity score between multimodal features.
It bridges the domain gap between visual $\mathbf{V} \in \mathbb{R}^{T\times D}$ and temporally localized textual $\mathbf{Q} \in \mathbb{R}^{L\times D}$ features. 
This is accomplished by contrastively learning the highly relevant encoded pseudo-query features $\mathbf{P} \in \mathbb{R}^{L_g\times D}$ that contain rich temporally global textual knowledge of the video for the respective moment interval, where $L_g$ denotes the total length of pseudo-quires.
Consequently, PIN assists the model in dynamically updating the weight in each learnable layer $f_{\theta}(\mathbf{V})$ and $f_{\sigma}(\mathbf{P})$, leading to a convergence of multi-modal features to align more closely within the feature space. To achieve this, we minimize the contrastive loss \cite{CLoss} with respect to mean averaged visual features $\tilde{\mathbf{V}} \in \mathbb{R}^{1 \times D}$ and pseudo-query features $\tilde{\mathbf{P}} \in \mathbb{R}^{1 \times D}$ in the batch $\mathcal{B}$,
\begin{equation}
    \mathcal{L}_{con} = \mathcal{L}_{q2v} + \mathcal{L}_{v2q},
\end{equation}
where,
\begin{equation}
    \mathcal{L}_{q2v} = -\frac{1}{\mathcal{B}}\sum_{i \in \mathcal{B}} \text{log} \frac{\text{exp}(\varepsilon \tilde{\mathbf{p}}^\top_i \tilde{\mathbf{v}}_i)}{\sum_{j \in \mathcal{B}} \text{exp}(\varepsilon \tilde{\mathbf{p}}^\top_i \tilde{\mathbf{v}}_j)},
\end{equation}
\begin{equation}
    \mathcal{L}_{v2q} = -\frac{1}{\mathcal{B}}\sum_{i \in \mathcal{B}} \text{log} \frac{\text{exp}(\varepsilon \tilde{\mathbf{v}}^\top_i \tilde{\mathbf{p}}_i)}{\sum_{j \in \mathcal{B}} \text{exp}(\varepsilon \tilde{\mathbf{v}}^\top_i \tilde{\mathbf{p}}_j)}.
\end{equation}
the $i$ and $j$ denotes $i$-th video and $j$-th pseudo-queries in the batch $\mathcal{B}$. $\varepsilon$ is the hyper-parameter that controls the strength of penalties for negative samples. 

To further bridge the domain gap of the multi-modal features, we randomly masked the M\% (e.g., M=30) of the object word in the pseudo queries. We utilized an additional projection network (MLP) to compute visual semantic tokens $\{\tilde{\mathbf{v}}_o\} \in \mathbb{R}^{N_{o} \times D}$ based on visual features $f_\delta(\tilde{\mathbf{V}})$, where $N_{o}$ denotes number of masked object words. These visual semantic tokens are integrated into the masked pseudo-query features $\mathbf{P} \in \mathbb{R}^{L_m \times D}$. We hypothesize that the visual object token can learn to represent the textual features if the video and pseudo query are aligned correspondingly. Furthermore, utilizing a masking strategy within the PIN framework also serves the purpose of mitigating the minimization of noisy (e.g., incorrect object words) information introduced by pseudo queries. With the refined visual query features, the model can compute higher similarity scores $\mathcal{S}$ in equation (2), leading to more precise predictions of target moments (e.g., higher IoU$@$m score as illustrated in Tables 2 and 3). 


\subsection{Prompt Guided Multi-modal Fusion}
Drawing inspiration from recent prompting-based transfer learning methods \cite{textPro, ProCL}, which preserve essential information for model enhancement, we apply a learnable prompt, PQ-prompt $\mathbf{\mathcal{P}} \in \mathbb{R}^{N \times D}$, to optimize the prompt pool containing the information temporally global textual knowledge, where $N$ denotes the length of the learnable prompt. 
In most cases, pseudo-queries do not exist during test time. Hence, at the inference stage, we directly use query $\mathbf{Q}$ which contains highly relevant information to retrieve $\mathbf{\mathcal{P}}$ that has temporally global textual knowledge stored in the prompt pool \cite{ProCL}. $\mathbf{\mathcal{P}}$ is then used to guide the pre-trained Prompt Guided Multi-modal Fusion (PGMF) module to predict a more accurate temporal moment.

Moreover, with PQ-prompt, the global textual knowledge and object semantic knowledge $\{\tilde{\mathbf{v}}_o\} \in \mathbb{R}^{N_{o} \times D}$ can be added to temporally localized ground truth query features $\mathbf{Q}$ to benefits alignment learning in PGMF at the training stage. 
To achieve this, we first append PQ-prompt $\mathbf{\mathcal{P}}$ with $\mathbf{P}$ and utilize a single-layer self-attention network to capture the token-level dependency within the pseudo-query features. Furthermore, PQ-prompt is concatenated to GT query features such that $\hat{\mathbf{Q}} = \text{Concat}(\mathbf{\mathcal{P}}; \mathbf{Q}) \in \mathbb{R}^{(L + N)\times D}$ and inject into PGMF module. The self-attention network is trained along with PIN that is optimized with the contrastive loss. 

\begin{figure}[!t]
\centering
\includegraphics[width=3.0in]{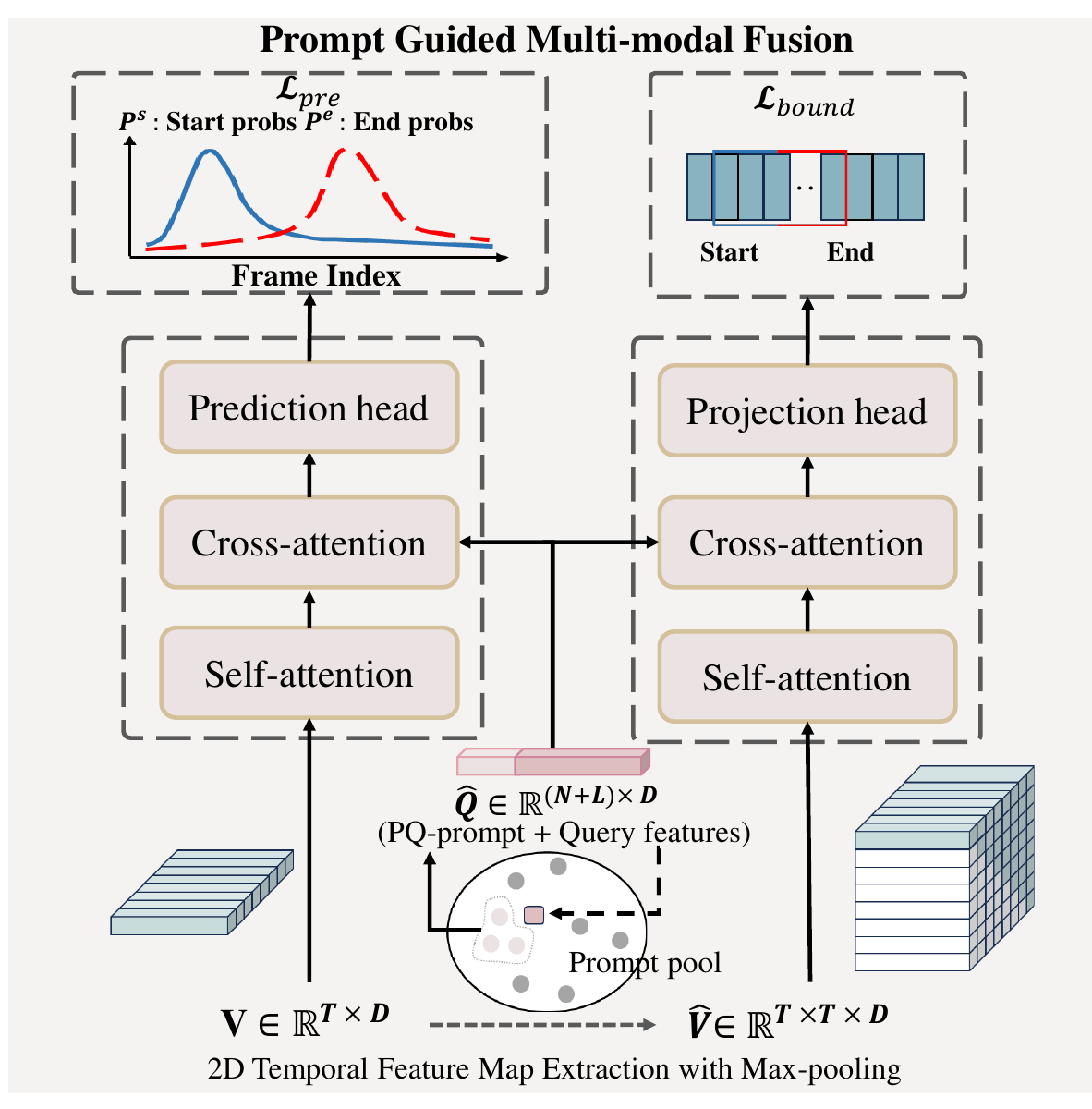}
\caption{The illustration of Prompt Guided Multi-modal Fusion module. 
}
\label{fig_3}
\end{figure}

Fig. \ref{fig_3} illustrated the learning of PGMF. We extract a 2D temporal feature map $\hat{\mathbf{V}} \in \mathbb{R} ^{T \times T \times D}$ using the Max-pool function with $\mathbf{V}$ by following the existing method \cite{2DTAN, EMB}. 
With PQ-prompt refined query features $\hat{\mathbf{Q}}$, refined visual features $\mathbf{V}$, we can formulate vision-language attention representation in PGMF with the following formula:
\begin{equation}
    \mathbf{V}^{'} = \mathcal{F}(\mathbf{V}) = \mathbf{V} + \text{FC}(\mathbf{V})\mathcal{S} \in \mathbb{R} ^ {T \times D}
\end{equation}
\begin{equation}
    \mathcal{S} = \text{Softmax}(\text{FC}(\mathbf{V})^\top\text{FC}(\mathbf{V}) \slash \sqrt{D})
\end{equation}
where FC denotes fully connected layer and $\mathcal{F}$ represents multihead-attention \cite{Transformer}. The PGMF is constructed by both self multihead-attention and cross multihead-attention of two modalities: 
    $\mathbf{Q}^{'} = \mathcal{F}(\hat{\mathbf{Q}}) \in \mathbb{R} ^ {(L+N) \times D}$,
    $\mathbf{V}^{''} = \mathcal{F}(\mathbf{V}^{'}, \mathbf{Q}^{'}) \in \mathbb{R} ^ {T \times D}$, and
    $\mathbf{Q}^{''} = \mathcal{F}(\mathbf{Q}^{'}, \mathbf{V}^{'}) \in \mathbb{R} ^ {(L+N) \times D}$ 
to explore the semantic alignment between video features and query features that are guided with PQ-prompt. 
The aligned features are further injected into the prediction head for the endpoint prediction. Similar to the existing method \cite{VSLNet}, we predict the frame-wise start and end boundaries probabilities with stacked two LSTM networks as:
\begin{equation}
    (P^s, P^e) = \text{Softmax}(\text{FC}(\text{LSTM}(\mathbf{V}^{'''}))) \in \mathbb{R} ^ {T}
\end{equation}
\begin{equation}
    \mathbf{V}^{'''} = \text{FC}([\mathbf{V''};
    \mathbf{V''}\odot \mathbf{Q''}\mathcal{A}_{r}^{\top};
    \mathbf{Q''} \odot \mathbf{V''}\mathcal{A}_{r}\mathcal{A}_{c}^{\top}])
\end{equation}
\begin{equation}
    \mathcal{A} = \text{FC}(\mathbf{V''})^\top\text{FC}(\mathbf{Q''}) \slash \sqrt{D}
\end{equation}
where [;] denotes concatenation operation, $\top$ represents the transpose function, and $\odot$ is the element-wise multiplication. $\mathcal{A}_{r}$ and $\mathcal{A}_{c}$ are computed based on the Softmax of the row- and column-wise normalization of $\mathcal{A}$, respectively. In addition, the highlighting strategy \cite{VSLNet} is included to align the PQ-prompt guided multi-modal features, where the highlighting score can be computed as $P^h = \sigma(\text{Conv1D}(\mathbf{V}^{'''})$. Overall, the endpoint prediction loss $\mathcal{L}_{pre}$ can be computed using the cross-entropy (CE) loss function, which is defined as:
\begin{equation}
    \mathcal{L}_{pre} = \frac{1}{2} [f_{CE}(P^s, Y^s) + f_{CE}(P^e, Y^e)] + \lambda_1 f_{CE}(P^h, Y^h)
\end{equation}
where $Y^e$, $Y^s$, and $Y^h$ are the ground truth of the start, end target moment labels, and highlighted foreground video moment \cite{VSLNet}, respectively.

For boundary optimization, we first flatten the 2D temporal feature map from $\hat{\mathbf{V}} \in \mathbb{R} ^{T \times T \times D}$ to $\hat{\mathbf{V}} \in \mathbb{R} ^{K \times D}$, then adopt function $\mathcal{F}$ to model the PQ-prompt guided cross-attention alignment between the modalities: 
$\hat{\mathbf{V}}^{'} = \mathcal{F}(\hat{\mathbf{V}}) \in \mathbb{R} ^ {(L+N) \times D}$,
$\hat{\mathbf{V}}^{''} = \mathcal{F}(\hat{\mathbf{V}}^{'}, \mathbf{Q}^{'}) \in \mathbb{R} ^ {K \times D}$, and
$\hat{\mathbf{Q}}^{''} = \mathcal{F}(\mathbf{Q}^{'}, \hat{\mathbf{V}}^{'}) \in \mathbb{R} ^ {(L+N) \times D}$. Similar to equation (9), we calculate $\hat{\mathbf{V}}^{'''} \in \mathbb{R} ^ {K \times D}$ base on $\hat{\mathbf{V}}^{''}$ and $\hat{\mathbf{Q}}^{''}$ in the projection head and reshape it back to 2D feature map $\hat{\mathbf{V}}^{'''} \in \mathbb{R} ^ {T \times T \times D}$. The matching score between a moment candidate with the PQ-prompt guided queried sentence can be formulated as:
\begin{equation}
    P_m = \sigma(\text{Conv2D}(\hat{\mathbf{V}}^{'''})) 
\end{equation}
the $\mathcal{L}_{bound}$ can be optimized as:
\begin{equation}
    \mathcal{L}_{bound} = \frac{1}{C}\sum^{C}_{m=1} y_m \text{log} p_m + (1-y_m)\text{log}(1-p_m)
\end{equation}
where C denotes valid moment candidates defined in 2DTAN \cite{2DTAN} and $y_m$ is the scaled IoU that serves as the supervision label. 

\noindent\textbf{Training stage.} The overall loss function of the proposed method is then formulated as follows:
\begin{equation}
    \mathcal{L} = \lambda_2 \mathcal{L}_{pre} + \lambda_3 \mathcal{L}_{bound} + \lambda_4 \mathcal{L}_{con}
\end{equation}
where $\lambda$ is the hyper-parameter to balance the loss terms.

\section{EXPERIMENT} 
\label{sec:pagestyle}
\subsection{Datasets and Evaluation Metrics}
In our experiments, the commonly used datasets Charades-STA and ActivityNet Captions are selected to evaluate the proposed method. Specifically, the Charades-STA provides 12408 and 3720 video-query pairs in its train and test subsets, respectively. The ActivityNet Captions contain 37417, 17505, and 17031 video-query pairs in its train, val\_1, and val\_2 subsets. Following the common selection in previous works, we evaluate all methods on the test subset of Charades-STA and the val\_2 subset of ActivityNet Captions. 

Intersection over Union (IoU) is adopted to evaluate the performance of all methods. We follow usual practices to predefine IoU thresholds $m$ = 0.3, 0.5, 0.7 to obtain a series of IoU@$m$ metrics, such that the percentage of predicted moments is considered correct if the IoU value is larger than $m$.

\begin{table}[t!]\setlength{\tabcolsep}{3.6pt}
  \centering
  \caption{Performance compared with the state-of-the-art models on Charades-STA and ActivityNet Captions dataset. All values are in percentages (\%), and higher is better. \textbf{Bold} indicates the best result, where the \underline{underline} number denotes the second best. It's worth noting that the pseudo queries are not included during testing for a fair comparison.}
    \begin{tabular}{l|ccc|ccc}
    \hline
    \multicolumn{1}{c|}{\multirow{3}[6]{*}{Method}} & \multicolumn{3}{c|}{Charades-STA} & \multicolumn{3}{c}{ActivityNet-Captions} \\
\cline{2-7}    \multicolumn{1}{c|}{} & \multicolumn{3}{c|}{IoU@m} & \multicolumn{3}{c}{IoU@m} \\
\cline{2-7}    \multicolumn{1}{c|}{} & 0.3   & 0.5   & \multicolumn{1}{c|}{0.7} & 0.3   & 0.5   & 0.7 \\
    \hline
    2DTAN \cite{2DTAN} &  -   & 39.81 & 23.31 & 59 .45 & 44.51 & 26.54 \\
    VSLNet \cite{VSLNet} & 64.30  & 47.31 & 30.19 & 63.16 & 43.22 & 26.16 \\
    BPNet \cite{BPNet} & 65.48 & 50.75 & 31.64 & 58.98 & 42.07 & 24.69 \\
    IVG \cite{IVG}  & 67.63 & 50.24 & 32.88 & 63.22 & 43.84 & \underline{27.10} \\
    HCL \cite{HCL}  & 71.56 & 57.39 & 34.41 & 63.05 & \textbf{45.82} & 26.79 \\
    EMB \cite{EMB} & 72.50  & 58.33 & 39.25 & \underline{64.13} & 44.81 & 26.07 \\
    EMTM \cite{EMTM} & \underline{72.70}  & 57.91 & \underline{39.80}  & 63.20  & 44.73 & 26.08 \\
    TCFN \cite{TCFN} & 68.58 & 51.77 & 29.95 & 56.81 & 40.58 & 24.73 \\
    UniVTG \cite{Univtg} & 72.63 & \textbf{60.19} & 38.55 &  -   &   -   & -\\
    \hline
    Ours  & \textbf{74.01} & \underline{59.76} & \textbf{41.43} & \textbf{65.38} & \underline{45.61} & \textbf{27.93} \\
    \hline
    \end{tabular} 
  \label{tab:addlabel}%
\end{table}%

\subsection{Implementation Details}
We adopted the  I3D video features \cite{I3D} used in the baseline model \cite{EMB} for the Charades-STA dataset \cite{Charades-STA}, used the C3D video features \cite{C3D} for ActivityNet Captions \cite{activitynet}, and 300D GloVe \cite{glove} to encode the embeddings of text inputs. The longer videos are downsampled to 128 frames by max-pooling, whereas the shorter videos are zero-padded. The hidden dimensions of all the layers are projected to 128D. The number of self-attention heads is set to 8, and random dropout is set to 0.2. The model is trained for 50 epochs with a batch size of 16. To optimize the training, we adopt an Adam optimizer under the settings of decaying learning rate ($5e^{-4}$) and gradient clipping (1.0). The hyper-parameter of losses was set to $\lambda_1$ = 5, $\lambda_2$ = $\lambda_3$ = 1 and $\lambda_4$ = 0.5 for both datasets empirically. The percentage of masking was set to M = 30\% for  Charades-STA (20\% for ActivityNet Captions). The number of Top-K pseudo-queries was empirically set to 3, and the length of prompt $N$ = 2 for both datasets. 

\subsection{Performance Evaluation}
\subsubsection{Quantitative Analysis}
We compare the proposed method with the SOTA temporal grounding methods, including 2DTAN \cite{2DTAN} (AAAI 2020), VSLNet \cite{VSLNet} (ACL 2020), BPNet \cite{BPNet} (AAAI 2021), IVG \cite{IVG} (CVPR 2021), HCL \cite{HCL} (MM 2022), EMB \cite{EMB} (ECCV 2022), EMTM \cite{TCFN} (2023), TCFN \cite{EMTM} (ICME 2023), and UniVTG \cite{Univtg} (ICCV 2023). 
Table 1 reports the results on the Charades-STA and ActivityNet-Captions datasets. We can see that the boundary established by our proposed method exhibits superior performance in most of the evaluation IoU scores among the existing SOTA methods. Specifically, compared to the recent best-performed baselines EMB, EMTM, and UniVTG, the absolute improvements of 1.31 to 1.38 on IoU = 0.3 and 1.63 to 2.18 on the extreme case for IoU = 0.7 can be observed for the Charades-STA dataset. Furthermore, for the ActivityNet-Captions dataset, the absolute improvements of 1.25 to 2.16 on IoU = 0.3 and 0.83 to 1.14 on the extreme case IoU = 0.7 can be observed as compared to the best baseline methods of  IVG, HCI, and EMB. These improvements have demonstrated the effectiveness of our proposed method for temporal grounding.

\begin{figure}[!t]
\centering
\includegraphics[width=3.3in]{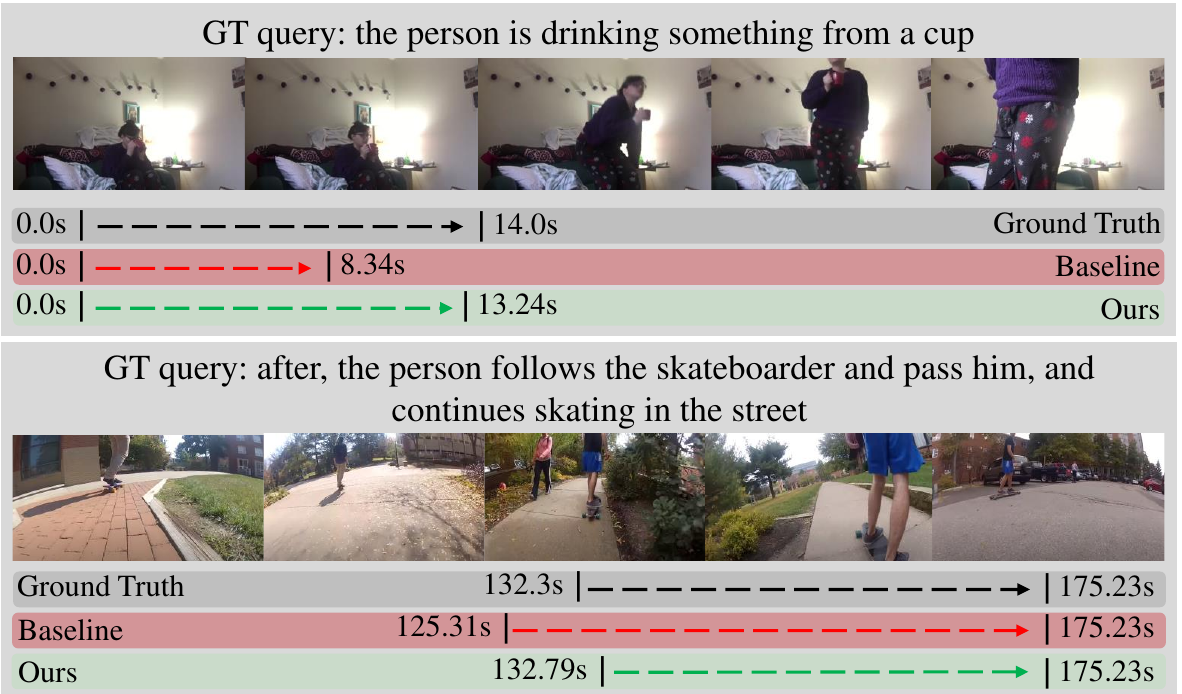}
\caption{Qualitative analysis with the baseline \cite{EMB} on Charades-STA (top) and ActivityNet (bottom) test set.}
\label{fig_4}
\end{figure}

\subsubsection{Qualitative Analysis} 
We showed two examples in Figure 4 from Charades-STA and ActivityNet datasets. From these two images, the localized moments predicted by our proposed method are closer to ground truth as compared to the baseline method \cite{EMB}. This implies that the proposed model can perform more satisfactorily with PIN and PQ-prompt guided multi-modal fusion module, which bridge the visual and language domain gaps for more precise temporal moments.

\subsubsection{Ablation Study}

In Tables \ref{tab:2}, we conducted ablation studies to verify the effectiveness of adopting PIN and PQ-prompt for Charades-STA and the ActivityNet Captions dataset. The cross indicates the baseline model that does not use PIN and PQ-prompt for training, whereas the tick denotes the component that has been adopted. We can observe that the model without using PIN and PQ-prompt will extremely damage the grounding performance. The proposed PIN shows improvements, demonstrating that it can improve multi-modalities alignment to enhance the grounding performance during testing. Furthermore, with the PQ-prompt to instruct the alignment learning in the multi-modal fusion module, the model has achieved significant improvement. The final model suppresses the baseline by 5.6 to 8.7\% on extreme cases for IoU = 0.7 for the Charades-
STA and the ActivityNet Captions dataset. These improvements demonstrate the effectiveness of the component that can predict boundaries more precisely.

\begin{table}[t!]\setlength{\tabcolsep}{2pt}
  \centering
  \caption{Impact of main components for proposed method}
    \begin{tabular}{c|c|ccc|ccc}
    \hline
    \multicolumn{1}{c|}{\multirow{3}[6]{*}{PIN}} & \multicolumn{1}{c|}{\multirow{3}[6]{*}{PQ-prompt}} & \multicolumn{3}{c|}{Charades-STA} & \multicolumn{3}{c}{ActivityNet-Captions} \\
\cline{3-8}          &       & \multicolumn{3}{c|}{IoU@m} & \multicolumn{3}{c}{IoU@m} \\
\cline{3-8}          &       & 0.3   & 0.5   & 0.7   & 0.3   & 0.5   & 0.7 \\
    \hline
    \xmark     &  \xmark & 72.66 & 58.31 & 39.22 & 63.83 & 43.72 & 25.70 \\
    \cmark     &  \xmark  & 73.31 & 59.13 & 40.38 & 64.40  & 44.84 & 26.47 \\
    \cmark     &  \cmark   & \textbf{74.01} & \textbf{59.76} & \textbf{41.43} & \textbf{65.38} & \textbf{45.61} & \textbf{27.93} \\
    \hline
    \end{tabular}%
  \label{tab:2}%
\end{table}%

\begin{table}[t!]\setlength{\tabcolsep}{2pt}
  \centering
  \caption{Impact of optimization loss for proposed method}
    \begin{tabular}{c|c|c|ccc|ccc}
    \hline
    \multicolumn{1}{c|}{\multirow{3}[5]{*}{Pre}} & \multicolumn{1}{c|}{\multirow{3}[5]{*}{Bound}} & \multicolumn{1}{c|}{\multirow{3}[5]{*}{Con}} & \multicolumn{3}{c|}{Charades-STA} & \multicolumn{3}{c}{ActivityNet-Captions} \\
\cline{4-9}          &       &       & \multicolumn{3}{c|}{IoU@m} & \multicolumn{3}{c}{IoU@m} \\
\cline{4-9}          &       &       & 0.3   & 0.5   & 0.7   & 0.3   & 0.5   & 0.7 \\
\hline
    \cmark      &   \xmark    &   \xmark    & 71.48 & 56.88 & 37.85 & 62.30 & 43.46 & 24.82 \\
    \cmark      &  \cmark     &   \xmark    & 72.45 & 58.29 & 39.19 & 64.21  & 44.21 & 26.16 \\
    \cmark      &   \cmark    &   \cmark    & \textbf{74.01} & \textbf{59.76} & \textbf{41.43} & \textbf{65.38} & \textbf{45.61} & \textbf{27.93} \\
    \hline
    \end{tabular}%
  \label{tab:3}%
\end{table}%

In Table \ref{tab:3}, we validate the effectiveness of the model when optimizing the model with different essential losses. We use Pre, Bound, and Con to denote the losses of $\mathcal{L}_{pre}$, $\mathcal{L}_{bound}$, and $\mathcal{L}_{con}$ in the table, respectively. We can observe that the additionally included $\mathcal{L}_{bound}$ and $\mathcal{L}_{con}$ can help to improve the performance significantly of 9.5\% to 12.5\% for IoU = 0.7 as compared to only includes the baseline $\mathcal{L}_{pre}$ loss \cite{VSLNet}, which showing the effectiveness to include PIN and Prompt Guided Multi-modal Fusion module with corresponding losses for better grounding.

\begin{table}[t!]\setlength{\tabcolsep}{4.8pt}
  \centering
  \caption{Experiment on the number of prompts that enhance the multi-modal learning in the Prompt Guided Multi-modal Fusion module.}
    \begin{tabular}{c|ccc|ccc}
    \hline
    \multirow{3}[6]{*}{Prompt} & \multicolumn{3}{c|}{Charades-STA} & \multicolumn{3}{c}{ActivityNet-Captions} \\
\cline{2-7}          & \multicolumn{3}{c|}{IoU@m} & \multicolumn{3}{c}{IoU@m} \\
\cline{2-7}          & 0.3   & 0.5   & 0.7   & 0.3   & 0.5   & 0.7 \\
    \hline
    0  & 73.31 & 59.13 & 40.38 & 64.40  & 44.84 & 26.47 \\
    1     & 73.66 & 59.61 & 41.07 & 65.19 & 45.22 & 27.22 \\
    2     & \textbf{74.01} & \textbf{59.76} & \textbf{41.43} & \textbf{65.38} & \textbf{45.61} & \textbf{27.93} \\
    3     & 73.46 & 59.62 & 40.94 & 65.13 & 45.02 & 27.03 \\
    4     &  73.45  &  59.54  &  40.67  &  64.78     &  44.93     & 26.84 \\
    \hline
    \end{tabular}%
  \label{tab:prompt} 
\end{table}%

\begin{table}[t!]\setlength{\tabcolsep}{4.8pt}
  \centering
  \caption{Experiment on masking percentage for semantic alignment in PIN.}
    \begin{tabular}{c|ccc|ccc}
    \hline
    \multirow{3}[6]{*}{Masking} & \multicolumn{3}{c|}{Charades-STA} & \multicolumn{3}{c}{ActivityNet-Captions} \\
\cline{2-7}          & \multicolumn{3}{c|}{IoU@m} & \multicolumn{3}{c}{IoU@m} \\
\cline{2-7}          & 0.3   & 0.5   & 0.7   & 0.3   & 0.5   & 0.7 \\
    \hline
    0\%  & 73.17 & 59.11 & 40.09 & 64.28 & 44.58 & 26.25 \\
    10\%  & 73.23 & 59.35 & 40.52 & 65.10 & 45.20 & 27.36 \\
    20\%  & 73.52 & 59.68 & 40.96 & \textbf{65.38} & \textbf{45.61} & \textbf{27.93} \\
    30\%  & \textbf{74.01} & \textbf{59.76} & \textbf{41.43} & 64.91 & 45.15 & 26.92 \\
    40\%  &   73.44    &  59.25     &   40.70    &   64.43    &   45.28    & 26.77 \\
    \hline
    \end{tabular}%
  \label{tab:masking} 
\end{table}%

Following the existing method \cite{ProCL, visualPro}, the learnable prompt emerges as a crucial hyperparameter that needs to be tuned for improved performance.
In TABLE \ref{tab:prompt}, we conduct ablation studies to verify the effectiveness of adopting various numbers ($N$) of learnable prompts to assist the multi-modal training for PGMF. We empirically set the number to $N$ = [0, 1, ..., 4]. We noticed that the model predicts inferior results when the learnable prompt is not included, whereas it was observed that the model achieves the best performance when we select the length of learnable prompts equal to 2. 

We further evaluate the performance of the model when masking different percentages of the object words in the pseudo-queries in TABLE \ref{tab:masking}. We can observe that the model performs optimally when 30\% of the object words are masked for Charades-STA and 20\% for the ActivityNet Captions dataset. 
For the Charades-STA dataset \cite{Charades-STA}, which primarily features relatively simple indoor scenes, we could mask out more object words since these scenes do not require as many additional object information to aid in model training compared to the ActivityNet Captions dataset \cite{activitynet}.  
When fewer object words are masked, there is a risk of introducing noise during the modality bridging process. 
Conversely, the higher masking percentage can lead to the alignment in less textual knowledge, resulting in a performance decline. 
Nevertheless, the method with masking outperformed the baseline without using a masking strategy.

\section{Conclusion}
In this work, we propose a Pseudo-query Intermediary Network (PIN) that bridges the domain gap between multi-modal features by contrastively aligning the visual features and pseudo-query features that contain comprehensive temporally global textual knowledge. Furthermore, we utilize the PQ-prompt that is enriched with temporally global textual knowledge to enhance alignment learning of prompt-guided multi-modal modules and achieve more precise target moment prediction. The extensive experiments on two datasets demonstrated the effectiveness of the proposed method.

\bibliographystyle{IEEEbib}
{\bibliography{icme2023template}}

\begin{thebibliography}{10}

\bibitem{2DTAN}
Songyang Zhang, Houwen Peng, Jianlong Fu, and Jiebo Luo,
\newblock ``Learning 2d temporal adjacent networks for moment localization with natural language,''
\newblock in {\em AAAI}, 2020.

\bibitem{HCL}
Bolin Zhang, Chao Yang, Bin Jiang, and Xiaokang Zhou,
\newblock ``Video moment retrieval with hierarchical contrastive learning,''
\newblock in {\em MM}, 2022.

\bibitem{VSLNet}
Hao Zhang, Aixin Sun, Wei Jing, and Joey~Tianyi Zhou,
\newblock ``Span-based localizing network for natural language video localization,''
\newblock in {\em ACL}, 2020.

\bibitem{TCFN}
Zezhong Lv and Bing Su,
\newblock ``Temporal-enhanced cross-modality fusion network for video sentence grounding,''
\newblock in {\em ICME}, 2023.

\bibitem{9356448}
Wenfei Yang, Tianzhu Zhang, Yongdong Zhang, and Feng Wu,
\newblock ``Local correspondence network for weakly supervised temporal sentence grounding,''
\newblock {\em IEEE Transactions on Image Processing}, 2021.

\bibitem{lan2023survey}
Xiaohan Lan, Yitian Yuan, Xin Wang, Zhi Wang, and Wenwu Zhu,
\newblock ``A survey on temporal sentence grounding in videos,''
\newblock {\em ACM Transactions on Multimedia Computing, Communications and Applications}, 2023.

\bibitem{Man}
Da~Zhang, Xiyang Dai, Xin Wang, Yuan-Fang Wang, and Larry~S Davis,
\newblock ``Man: Moment alignment network for natural language moment retrieval via iterative graph adjustment,''
\newblock in {\em CVPR}, 2019.

\bibitem{PGA}
Daizong Liu, Xiaoye Qu, and Pan Zhou,
\newblock ``Progressively guide to attend: An iterative alignment framework for temporal sentence grounding,''
\newblock in {\em EMNLP}, 2021.

\bibitem{CPN}
Kun Li, Dan Guo, and Meng Wang,
\newblock ``Proposal-free video grounding with contextual pyramid network,''
\newblock in {\em AAAI}, 2021.

\bibitem{GIC}
Chen Cai, Suchen Wang, Kim-Hui Yap, and Yi~Wang,
\newblock ``Top-down framework for weakly-supervised grounded image captioning,''
\newblock {\em Knowledge-Based Systems}, vol. 287, pp. 111433, 2024.

\bibitem{AMR}
Meng Liu, Xiang Wang, Liqiang Nie, Xiangnan He, Baoquan Chen, and Tat-Seng Chua,
\newblock ``Attentive moment retrieval in videos,''
\newblock in {\em SIGIR}, 2018.

\bibitem{zhang2023temporal}
Hao Zhang, Aixin Sun, Wei Jing, and Joey~Tianyi Zhou,
\newblock ``Temporal sentence grounding in videos: A survey and future directions,''
\newblock {\em IEEE Transactions on Pattern Analysis and Machine Intelligence}, 2023.

\bibitem{EMB}
Jiabo Huang, Hailin Jin, Shaogang Gong, and Yang Liu,
\newblock ``Video activity localisation with uncertainties in temporal boundary,''
\newblock in {\em ECCV}, 2022.

\bibitem{IVG}
Guoshun Nan, Rui Qiao, Yao Xiao, Jun Liu, Sicong Leng, Hao Zhang, and Wei Lu,
\newblock ``Interventional video grounding with dual contrastive learning,''
\newblock in {\em CVPR}, 2021.

\bibitem{ACC}
Chen Cai, Kim-Hui Yap, and Suchen Wang,
\newblock ``Attribute conditioned fashion image captioning,''
\newblock in {\em 2022 IEEE International Conference on Image Processing (ICIP)}, 2022, pp. 1921--1925.

\bibitem{PSL}
Minghang Zheng, Shaogang Gong, Hailin Jin, Yuxin Peng, and Yang Liu,
\newblock ``Generating structured pseudo labels for noise-resistant zero-shot video sentence localization,''
\newblock in {\em ACL}.

\bibitem{ProCL}
Zifeng Wang, Zizhao Zhang, Chen-Yu Lee, Han Zhang, Ruoxi Sun, Xiaoqi Ren, Guolong Su, Vincent Perot, Jennifer Dy, and Tomas Pfister,
\newblock ``Learning to prompt for continual learning,''
\newblock in {\em CVPR}, 2022.

\bibitem{textPro}
Brian Lester, Rami Al-Rfou, and Noah Constant,
\newblock ``The power of scale for parameter-efficient prompt tuning,''
\newblock in {\em EMNLP}, 2021.

\bibitem{visualPro}
Menglin Jia, Luming Tang, Bor-Chun Chen, Claire Cardie, Serge Belongie, Bharath Hariharan, and Ser-Nam Lim,
\newblock ``Visual prompt tuning,''
\newblock in {\em ECCV}, 2022.

\bibitem{Transformer}
Ashish Vaswani, Noam Shazeer, Niki Parmar, Jakob Uszkoreit, Llion Jones, Aidan~N Gomez, {\L}ukasz Kaiser, and Illia Polosukhin,
\newblock ``Attention is all you need,''
\newblock {\em NeurIPS}, 2017.

\bibitem{blip2}
Junnan Li, Dongxu Li, Silvio Savarese, and Steven Hoi,
\newblock ``Blip-2: Bootstrapping language-image pre-training with frozen image encoders and large language models,''
\newblock {\em arXiv preprint arXiv:2301.12597}, 2023.

\bibitem{glove}
Jeffrey Pennington, Richard Socher, and Christopher Manning,
\newblock ``{G}lo{V}e: Global vectors for word representation,''
\newblock in {\em EMNLP}, 2014.

\bibitem{GPT2}
Alec Radford, Jeffrey Wu, Rewon Child, David Luan, Dario Amodei, Ilya Sutskever, et~al.,
\newblock ``Language models are unsupervised multitask learners,''
\newblock {\em OpenAI blog}, 2019.

\bibitem{CLoss}
Kihyuk Sohn,
\newblock ``Improved deep metric learning with multi-class n-pair loss objective,''
\newblock {\em NeurIPS}, 2016.

\bibitem{BPNet}
Shaoning Xiao, Long Chen, Songyang Zhang, Wei Ji, Jian Shao, Lu~Ye, and Jun Xiao,
\newblock ``Boundary proposal network for two-stage natural language video localization,''
\newblock in {\em AAAI}, 2021.

\bibitem{EMTM}
Renjie Liang, Yiming Yang, Hui Lu, and Li~Li,
\newblock ``Efficient temporal sentence grounding in videos with multi-teacher knowledge distillation,''
\newblock {\em arXiv preprint arXiv:2308.03725}, 2023.

\bibitem{Univtg}
Kevin~Qinghong Lin, Pengchuan Zhang, Joya Chen, Shraman Pramanick, Difei Gao, Alex~Jinpeng Wang, Rui Yan, and Mike~Zheng Shou,
\newblock ``Univtg: Towards unified video-language temporal grounding,''
\newblock in {\em ICCV}, 2023.

\bibitem{I3D}
Joao Carreira and Andrew Zisserman,
\newblock ``Quo vadis, action recognition? a new model and the kinetics dataset,''
\newblock in {\em CVPR}, 2017.

\bibitem{Charades-STA}
Jiyang Gao, Chen Sun, Zhenheng Yang, and Ram Nevatia,
\newblock ``Tall: Temporal activity localization via language query,''
\newblock in {\em Proceedings of the IEEE international conference on computer vision}, 2017, pp. 5267--5275.

\bibitem{C3D}
Du~Tran, Lubomir Bourdev, Rob Fergus, Lorenzo Torresani, and Manohar Paluri,
\newblock ``Learning spatiotemporal features with 3d convolutional networks,''
\newblock in {\em ICCV}, 2015.

\bibitem{activitynet}
Fabian Caba~Heilbron, Victor Escorcia, Bernard Ghanem, and Juan Carlos~Niebles,
\newblock ``Activitynet: A large-scale video benchmark for human activity understanding,''
\newblock in {\em Proceedings of the ieee conference on computer vision and pattern recognition}, 2015, pp. 961--970.

\end{thebibliography}

\end{document}